\documentclass[letterpaper, 10 pt, journal]{ieeeconf} 

\usepackage{graphics} 
\usepackage{graphicx}   
\usepackage{amsmath} 
\usepackage{amssymb}  
\usepackage{mathtools, cuted}
\usepackage{interval}
\usepackage{nicefrac}
\usepackage[noadjust]{cite}
\usepackage{breqn}
\usepackage{color, colortbl}
\usepackage{derivative}
\usepackage{siunitx}
\usepackage{multirow}
\usepackage{cleveref}

\renewcommand{\vec}[1]{\mathbf{\boldsymbol{#1}}}

\providecommand{\R}{\ensuremath \mathbb{R}}

\newcommand{\de}{{\mathrel{\mathop:}=}}
\newcommand{\etal}{\textit{et al}. }
\newcommand{\ie}{\textit{i}.\textit{e}., }

\usepackage[dvipsnames]{xcolor}

\title{Energetic Analysis on the Optimal Bounding Gaits \\ of Quadrupedal Robots}

\author{Yasser G. Alqaham, Jing Cheng, Zhenyu Gan
\thanks{All authors are with the Department of Mechanical and Aerospace Engineering, Syracuse University, Syracuse, NY 13244 \texttt{\{ygalqaha, jcheng13, zgan02\}@syr.edu}.}
\thanks{This work was supported by a startup fund from the Syracuse University.}
}

\begin{document}

\maketitle

\begin{abstract}
It is often overlooked by roboticists when designing locomotion controllers for their legged machines, that energy consumption plays an important role in selecting the best gaits for locomotion at high speeds or over long distances.
The purpose of this study is to examine four similar asymmetrical quadrupedal gaits that are frequently observed in legged animals in nature.
To understand how a specific footfall pattern will change the energetics of a legged system, we first developed a full body model of a quadrupedal robot called A1.
And for each gait we created a hybrid system with desired footfall sequence and rigid impacts.
In order to find the most energy efficient gait, we used optimal control methods to formulate the problem as a trajectory optimization problem with proper constraints and objective function.
This problem was implemented and solved in a nonlinear programming framework called FROST.
Based on the optimized trajectories for each gait, we investigated the values of cost of transport and the work done by all joints.
Moreover, we analyzed the exchange of angular momentum in different components of the system during the whole stride cycle. 
According to the simulation results, bounding with two flight phases is likely to be the most energy efficient gait for A1 across a wide range of speed.
\end{abstract}


\section{Introduction}

Animals in nature coordinate their legs and locomote in a variety of ways, depending on their speed and the type of terrain.
The rhythmic limbs movement used for locomotion in animals is usually referred to as \emph{gait}. 
Researchers such as Milton Hildebrand investigated hundreds of quadrupedal gaits of vertebrates and classified them into two groups based on whether their footfalls of forelimb and hindlimb pairs are evenly spaced in time, namely, symmetrical gaits and asymmetrical gaits \cite{Hildebrand1965, hildebrand1977analysis}. 
Although symmetrical gaits, such as walking and trotting, have been extensively researched at relatively low speeds, asymmetrical gaits, such as bounding, half-bounding, and galloping, have received considerably less attention.
These asymmetrical gaits are commonly used by mammals at high speeds and frequently include aerial phases with notable body rotations \cite{hildebrand1977analysis, dunham2020asymmetrical}.
The bounding gait for example is usually defined as the one in which the front or rear leg pairs are always synchronized.
Depending on the pattern of footfall, there are several types of bounding gait across species.
They can be distinguished by the whole-body aerial phases (suspensions) within one stride: some animals, such as, giraffes are always supported by at least one stance leg and have no periods of suspension; camels and deer only have one interval of suspension in each cycle; and carnivores like cheetahs havE two suspensions in each cycle \cite{hildebrand1977analysis}.
In addition, during the aerial phase, the legs of camels are gathered under the body, while those of deer are extended forward and backward.
In order to build legged robots that are able to move at higher speeds, roboticists are highly motivated to test gait patterns that are similar to those found in nature. 
It has been more than thirty years since the first attempts were made to mimic bounding motion on legged systems. 
Murphy \etal designed a three-part controller to stabilize a planner articulated robot using a bounding gait in simulation \cite{Murphy1985}.
A three dimensional quadrupedal robot was built and tested using similar control methodologies on the bounding gait with two suspensions \cite{raibert1986running, raibert1986legged, raibert1990trotting}.
Later on, Poulakakis \etal developed several feedback control laws which resulted in bounding gaits with different footfall patterns on a quadrupedal robot called Scout II \cite{Poulakakis2003, poulakakis2006dynamics}. 
The fastest land speed record of legged robots was set by the ``Cheetah" robot, built by Boston Dynamics, bounding at speeds up to 18 miles per hour (mph) in 2012 \cite{Cheetah2012}.
More recently, Park \etal were able to incorporate the bounding gait into the MIT Cheetah II robot using model predictive control to command the robot to leap over obstacles of varying heights \cite{park2017high, park2021jumping}.


\begin{figure}[tbp]
\centering
\includegraphics[width=1\columnwidth]{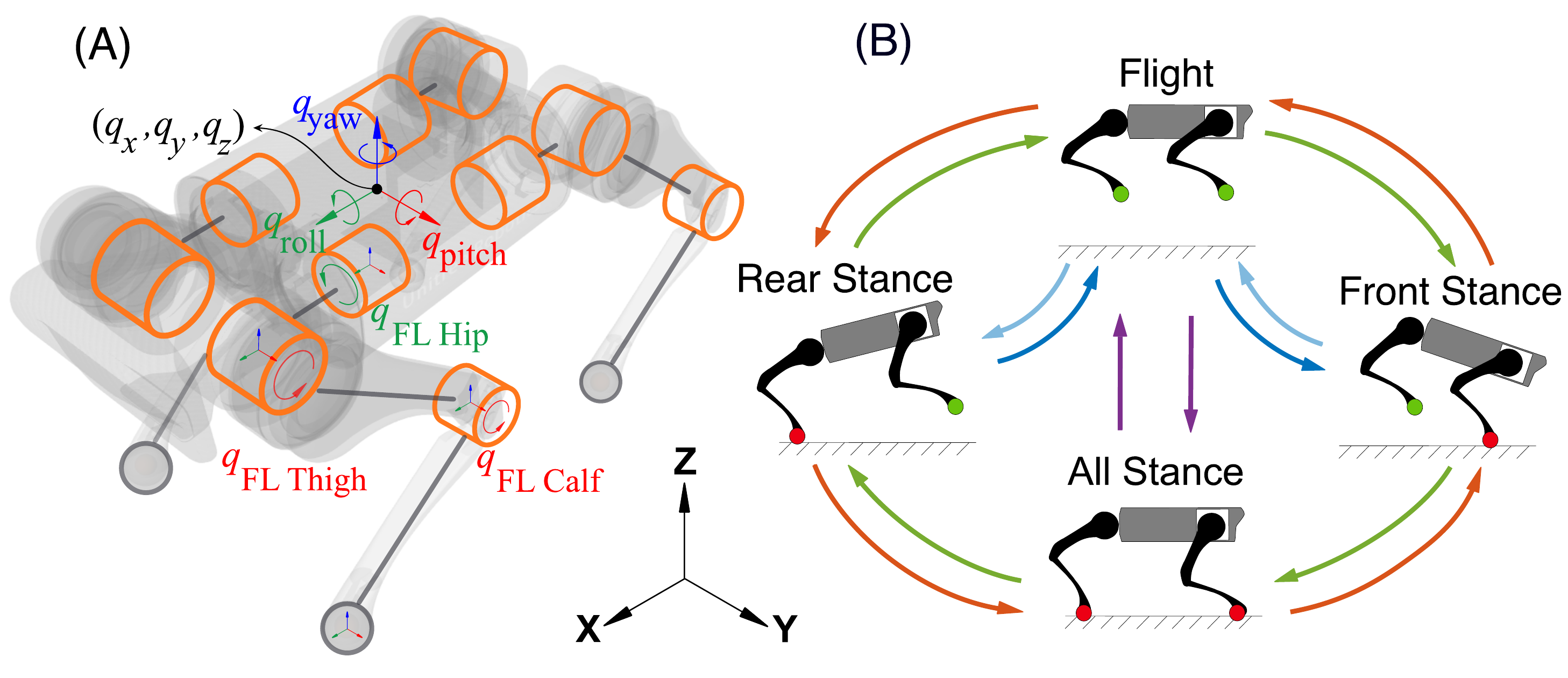}
\caption[Kinematic Model]{Figure (A) shows the kinematic model of A1 robot along with configuration variables. Each leg has three joints namely hip, thigh, and calf. Figure (B) demonstrates the footfall patterns for pronking and three bounding gaits.} 
\label{fig:Kinematics Diagram}
\vspace{-2mm}
\end{figure}
However, most of the aforementioned studies only focused on robot design and how to control the robot to follow a specific bounding pattern. 
Only a limited amount of attention was given to the differences among the several feasible bounding gaits, and how these differences are influenced by the specific structural design of the robot.
Animals in nature, on the other hand, have been hypothesized to reduce their energy expenditure when covering long distances \cite{hoyt1981gait}.
A similar concept called the dimensionless net metabolic Cost of Transport (COT) was widely adopted by roboticists to measure the energy efficiency of legged locomotion.
It is usually defined as the amount of energy consumed by a unit body mass to travel for a unit distance during one stride cycle \cite{donelan2004mechanical, bhounsule2014low, xi2014optimal, xi2016selecting, yesilevskiy2018spine}.
In this work, we seek to examine the fundamental differences between the bounding gaits with various suspensions as shown in Fig.~\ref{fig:Kinematics Diagram}(B) and try to identify how a specific design of the robot can impact the energy expenditures while moving at various speeds.
For this purpose, we built a detailed full body model of a popular quadrupedal robot called A1 from Unitree Robotics \cite{Unitree}.
Moreover, for each of the bounding gaits mentioned previously we created hybrid models with rigid impacts \cite[Chapter~3]{westervelt2018feedback}.
After that, we utilized the trajectory optimization technique in FROST framework implemented by Hereid \etal \cite{hereid2017frost, hereid2019rapid} to find the optimal solutions at each preferred speed.
In the end, we compared the values of COT, the positive and negative work performed in each joint, and the angular momentum exchanges in different components of the A1 robot to provide detailed information on the fundamental difference in all three bounding gaits.

\section{Methods}
\label{sec:methods}

This section provides a detailed description of the model of the quadrupedal robot A1, its hybrid dynamics, and the types of gaits we are interested in \ie pronking and bounding.
To generate realistic and efficient periodic motion, we solve a nonlinear trajectory optimization problem with the proposed footfall patterns \cite{gan2018}.
The details of the selection of cost functions and implementation of physical constraints are also provided in the this section.

\subsection{Robot Coordinates and Floating-Base Dynamic Model}
The quadrupedal A1 robot shown in Fig.~\ref{fig:Kinematics Diagram}(A) is designed and built by Unitree Robotics.
Without additional peripheral sensors, the total mass of the robot is about 12.45 kg and each leg has three electric motors to actuate the hip, thigh, and calf joints, respectively.

Assuming all moving parts on the robot are rigid bodies with distributed mass, the floating-base model (FBM) of A1 has in total 18 degrees of freedom (DOF) as illustrated in Fig.~\ref{fig:Kinematics Diagram}(A). We use variables $\vec{q_{\text{body}}} \de \left[q_x,q_y,q_z,q_{\text{yaw}},q_{\text{pitch}},q_{\text{roll}} \right]^\intercal$ to represent the Cartesian position of the torso's geometrical center in the inertial coordinate frame and the torso’s intrinsic Euler angles in the $z{\text -}y{\text -}x$ order, respectively.
For $i$th leg, the joint vector $\vec{q}_{i} \de \left[q_{i,\text{hip}},q_{i,\text{thigh}},q_{i,\text{calf}}\right]^\intercal$ refers to the relative angles of the joints measured in its own body coordinate frames.
The index $ i \in {\{\text{FR}, \text{FL}, \text{RR}, \text{RL}\}}$ stands for the front right, the front left, the rear right, and the rear left legs.
The leg coordinates are $ \vec{q}_{\text{leg}} \de \left[\vec{q}_{\text{FR}}^\intercal, \vec{q}_{\text{FL}}^\intercal, \vec{q}_{\text{RR}}^\intercal, \vec{q}_{\text{RL}}^\intercal \right] ^\intercal  \in\R^{12}$.
The generalized coordinates of the FBM are aggregated in a single vector and defined as follows:
\begin{equation}
\label{eq:configuration variables}
       \vec{q} \de \left[\vec{q}_{\text{body}}^\intercal, \vec{q}_{\text{leg}}^\intercal \right] ^\intercal
\end{equation}
%
%

By applying Euler-Lagrange equation, the equations of motion of the A1 robot can be expressed in the following form:
\begin{equation}
\label{eq:EOM}
   \operatorname{\vec{M}}(\vec{q})\ddot{\vec{q}} + \operatorname{\vec{C}}(\vec{q},\dot{\vec{q}}) + \operatorname{\vec{G}}(\vec{q}) = \vec{S}\vec{\tau}  + \vec{J}^\intercal(\vec{q})\vec{\lambda} .
\end{equation}
where $\operatorname{\vec{M}}(\vec{q})\in\R^{18 \times 18}$ is the inertia matrix; $\operatorname{\vec{C}}(\vec{q},\dot{\vec{q}})\in\R^{18}$ is the vector consisting of Coriolis forces and centrifugal forces;
and $\operatorname{\vec{G}}(\vec{q})\in\R^{18}$ is the vector of gravitational forces. 
$\vec{\tau} \in\R^{12}$ denotes the vector of joints motors torques and $\vec{S} = \left[\vec{0}_{6 \times 12}; \vec{I}_{12 \times 12}\right]$ is the selection matrix which contains a zero submatrix and an identity submatrix to assign motor torques to the corresponding joints.
The last term $\vec{J}^\intercal(\vec{q})\vec{\lambda}$ is the system external forces mapped to the joints space where $\vec{J}(\vec{q})\in\R^{3 \times 18}$ is the Jacobian matrix (explained in the next section) and $\vec{\lambda}\in\R^{3}$ is the external forces vector. 

\subsection{Hybrid Model for Pronking and Bounding}
During pronking and bounding gaits, the robot passes through several phases in which different pairs of legs alternate between stance (legs in contact with the ground and not slipping) and flight (legs moving freely above the ground).
We employ four domains in order to determine periodic gaits: all stance, front stance, rear stance, and flight.
All four feet of robot A1 are made of rubber balls measuring 0.02 meters in diameter; therefore, we model the stance foot constraints as point contacts and enforce the foot accelerations to be zeros.
When the $i$th leg is in stance, the Cartesian position of the foot remains stationary.
This imposes the following holonomic constraints: 
\begin{align}
        \vec{c}_i - \vec{g}_{i}\left(\vec{q}\right) = \vec{0} , \quad \notag
        \vec{J}_i(\vec{q}) \dot{\vec{q}} = \vec{0}, \\
        \vec{J}_i(\vec{q}) \ddot{\vec{q}} + \vec{\dot{J}}_i(\vec{q}, \dot{\vec{q}}) \dot{\vec{q}} = \vec{0}, 
        \label{eq:cv}
\end{align}
where $\vec{c}_i$ is the desired foot locations in Cartesian coordinate; $\vec{g}_i \de \left[ p^x_i(\vec{q}), \: p^y_i(\vec{q}), \: p^z_i(\vec{q}) \right]^\intercal$ denotes the set of forward kinematic equations of the foot position in the inertial frame; $\vec{J}_i \mathrel{\mathop:}=  \pdv{\vec{g}_i(\vec{q})}{\vec{q}}$ is the corresponding Jacobian matrix of the foot contact.
The corresponding ground reaction forces $\vec{\lambda}_i \de \left[ f^x_i, \: f^y_i, \: f^z_i \right]^\intercal$ have the same dimension of $\vec{g}_i$.

These four domains therefore are distinguished by the different numbers of holonomic constraints and the presence of external forces as follows:
\subsubsection{All Stance $\mathcal{F}_{A}$}
in this domain, the above holonomic constraints are active for all four legs where $A = {\{\text{FR}, \text{FL}, \text{RR}, \text{RL}\}}$ and $\vec{J}_{A} ^\intercal = \left[ \vec{J}_{\text{FR}} ^\intercal, \vec{J}_{\text{FL}} ^\intercal, \vec{J}_{\text{RR}} ^\intercal, \vec{J}_{\text{RL}} ^\intercal \right]$.We collect the GRFs of all four legs in the same order and denote them as $\vec{\lambda}_{A} \in \R^{12}$.

\subsubsection{Front Stance $\mathcal{F}_{F}$} 
in this domain, the holonomic constraints are only enforced on the front leg pairs where $F = {\{\text{FR}, \text{FL}\}}$ and $\vec{J}_{F} ^\intercal = \left[ \vec{J}_{\text{FR}} ^\intercal, \vec{J}_{\text{FL}} ^\intercal \right]$. The constrained EOM is similar to \eqref{eq:DAE} with the corresponding Jacobian matrix.
\subsubsection{Rear Stance $\mathcal{F}_{R}$}
similar to front stance, the holonomic constraints are only enforced on the rear leg pairs where $R = {\{\text{RR}, \text{RL}\}}$ and $\vec{J}_{R} ^\intercal = \left[ \vec{J}_{\text{RR}} ^\intercal, \vec{J}_{\text{RL}} ^\intercal \right]$.

\subsubsection{Flight $\mathcal{F}_{T}$}
the robot follows a ballistic motion in flight. All legs are above the ground, and there is no ground reaction forces $\vec{\lambda}$. 

By combining \eqref{eq:EOM} and the corresponding holonomic constraints of each domain listed above, we obtain the following differential-algebraic equations for a domain $\Gamma_{*}$ ($*$ indicates one of the cases above) with the different number of feet that are in contact with the ground:
\begin{equation}
\label{eq:DAE} 
\resizebox{.9 \linewidth}{!}{$  
    \begin{bmatrix}
           \operatorname{\vec{M}}(\vec{q}) & -\vec{J}_{\mathcal{*}}^\intercal(\vec{q})  \\
           \vec{J}_{\mathcal{*}}(\vec{q}) &  \vec{0}
    \end{bmatrix}
     \begin{bmatrix}
           \ddot{\vec{q}} \\
           \vec{\lambda}_{\mathcal{*}}
    \end{bmatrix}
    = \begin{bmatrix}
           \vec{S}\vec{\tau} - \operatorname{\vec{C}}(\vec{q},\dot{\vec{q}})\dot{\vec{q}} - \operatorname{\vec{G}}(\vec{q})  \\
           -\vec{\dot{J}}_{*}(\vec{q}, \dot{\vec{q}})\dot{\vec{q}}
    \end{bmatrix}.  $}    
\end{equation}

For each leg, the transition from flight phase to stance phase is captured by the vertical position of the foot crossing zero from positive direction.
The touch-down (TD) transition set therefore can be written as:
\begin{equation}
\label{eq:TD}
  \mathcal{S}_{i}^{\text{TD}} =  \{ \left( \vec{q},\dot{\vec{q}} \right) \in \mathcal{TQ} \; | \; p^z_i = 0, \; \dot{p}^z_i <0   \} .
\end{equation}
Whenever a leg transitions from stance phase to swing phase, the vertical ground contact force $f^z_i$ decreases to zero.
We can define the lift-off (LO) transition set as follows:
\begin{equation}
\label{eq:LO}
  \mathcal{S}_{i}^{\text{LO}} =  \{ \left( \vec{q},\dot{\vec{q}} \right) \in \mathcal{TQ} \; | \; f^z_i = 0, \; \dot{f}^z_i <0   \} .
\end{equation}
We use the rigid impact model where the post impact velocities $\dot{\vec{q}}^+$ is a linear fumction of the pre-impact velocity $\dot{\vec{q}}^-$. The reset map $\Delta_{j \rightarrow k}$ $\left(j,k \in \{\mathcal{F}_{A}, \mathcal{F}_{F}, \mathcal{F}_{R}, \mathcal{F}_{T}\}, j \not= k \right)$ can be easily calculated according to \cite[Chapter~3]{westervelt2018feedback}.

\begin{figure}[tbp]
\centering
\includegraphics[width=1\columnwidth]{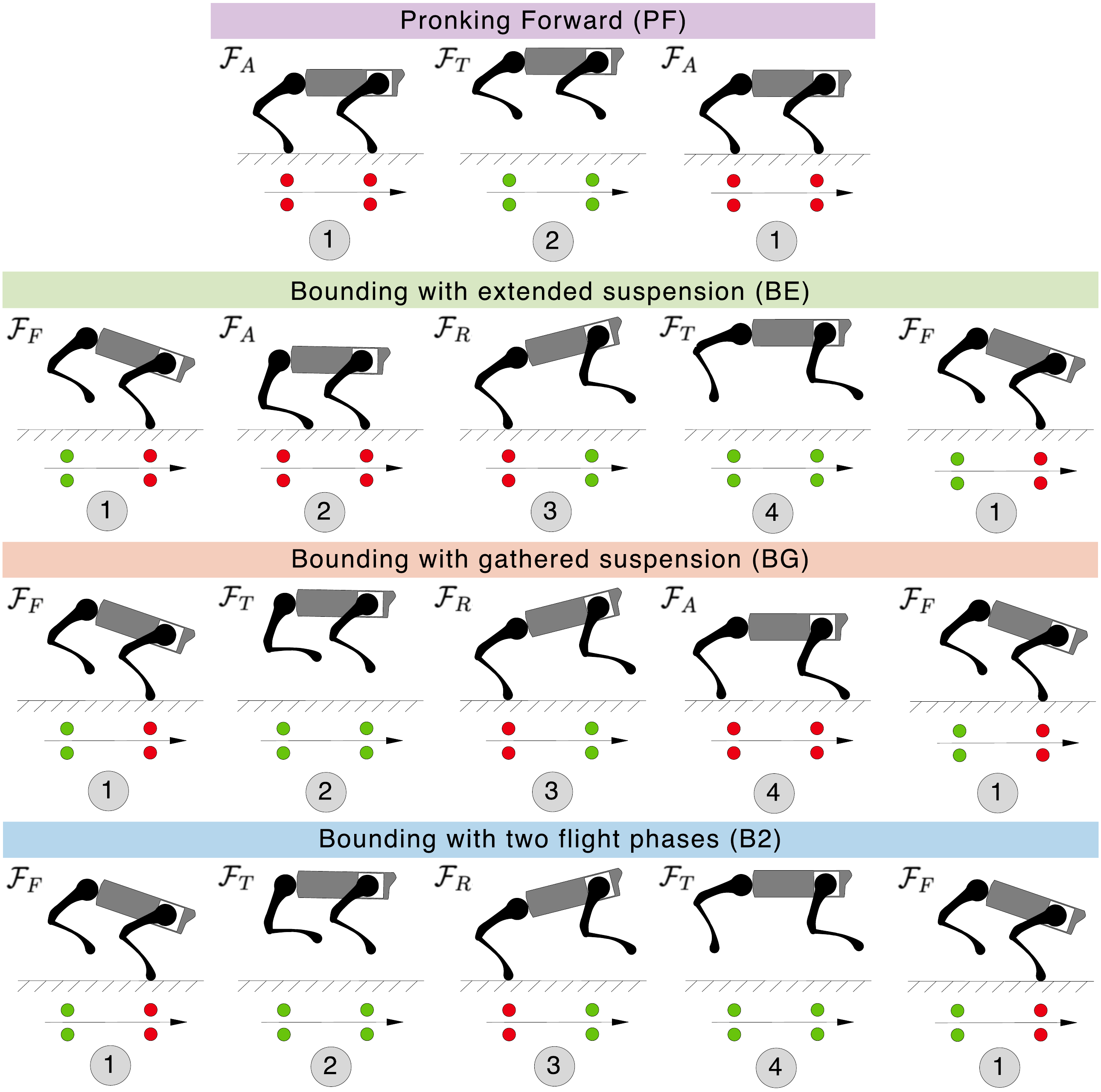}
\caption[Iterative learning control algorithm]{In all bounding gaits, the robot motion cycles through four phases of flight, front-leg stance, rear-leg stance, and all-leg stance each with different sequence. In B2 bounding gait, it is the same four phases but instead of all-leg stance we have another flight phase. The pronking gait (PF) has only two phases: all-leg stance and flight.}
\label{fig:FootFall Diagram}
\vspace{-2mm}
\end{figure}

In this work, we consider four asymmetrical gaits as depicted in Fig.~\ref{fig:FootFall Diagram}: pronking forward (PF), bounding with extended suspensions (BE), bounding with gathered suspensions (BG), and bounding with two flight phases (B2) \cite{gan2018}.
In a pronking gait, all four legs move simultaneously during the entire stride cycle.
The overall hybrid model for the pronking gait $\Sigma_{\text{PF}}$ is given as follows:
\begin{equation} 
\resizebox{.91 \linewidth}{!}{$  \hspace{-0.2cm} \Sigma_{\text{PF}}: \left\{ 
    \begin{array}{llrr}
         \mathcal{F}_{A}, 
         &\hspace{-0.2cm} \left( \vec{q},\dot{\vec{q}} \right) \not \in \mathcal{S}^{\text{LO}}_{A}; 
         &\dot{\vec{q}}^+ = \Delta_{\mathcal{F}_{A} \rightarrow \mathcal{F}_{T}} \dot{\vec{q}}^-,
         &\hspace{-0.2cm}\left( \vec{q},\dot{\vec{q}} \right) \in \mathcal{S}^{\text{LO}}_{A}; \\ 
         \mathcal{F}_{T}, 
         &\hspace{-0.2cm}\left( \vec{q},\dot{\vec{q}} \right) \not \in \mathcal{S}^{\text{TD}}_{A};
         &\dot{\vec{q}}^+ = \Delta_{\mathcal{F}_{T} \rightarrow \mathcal{F}_{A}} \dot{\vec{q}}^-,  
         &\hspace{-0.2cm}\left( \vec{q},\dot{\vec{q}} \right)  \in \mathcal{S}^{\text{TD}}_{A}; \\
    \end{array}
\right .$}    
\label{eq:HybridModelPronk}
\end{equation}   

The three bounding gaits share the same number of domains in each period, and each leg pair (front or rear) move in a synchronized fashion. The sequence of events as well as footfall patterns in all three bounding gaits are distinct. 
The BE and BG bounding gaits are featured by a single aerial phase. The major difference between them is during the flight phase when the swing legs are extended outward in BE (as seen in the second row of Fig.~\ref{fig:FootFall Diagram} whereas the swing legs are gathered toward the center in BG. Bounding B2 is distinguished by having two flight phases where the legs alternate between swinging inward and swinging outward. The hybrid models $\Sigma_{\text{BE}}$, $\Sigma_{\text{BG}}$, and $\Sigma_{\text{B2}}$ are summarized below:

\begin{equation} 
\label{eq:HybridModelBE}
\resizebox{.89 \linewidth}{!}{$ \hspace{-0.2cm}\Sigma_{\text{BE}}:   \left\{ 
    \begin{array}{rrll}
         \mathcal{F}_F,
         &\hspace{-0.2cm}\left( \vec{q},\dot{\vec{q}} \right) \not \in \mathcal{S}^{\text{TD}}_{R};
         &\dot{\vec{q}}^+ = \Delta_{\mathcal{F}_{F} \rightarrow \mathcal{F}_{A}} \dot{\vec{q}}^-,
         &\hspace{-0.2cm}\left( \vec{q},\dot{\vec{q}} \right) \in \mathcal{S}^{\text{TD}}_{R}; \\
         \mathcal{F}_A,
         &\hspace{-0.2cm}\left( \vec{q},\dot{\vec{q}} \right) \not \in \mathcal{S}^{\text{LO}}_{F};
         &\dot{\vec{q}}^+ = \Delta_{\mathcal{F}_{A} \rightarrow \mathcal{F}_{R}} \dot{\vec{q}}^-,
         &\hspace{-0.2cm}\left( \vec{q},\dot{\vec{q}} \right)  \in \mathcal{S}^{\text{LO}}_{F}; \\
         \mathcal{F}_R,
         &\hspace{-0.2cm}\left( \vec{q},\dot{\vec{q}} \right) \not \in \mathcal{S}^{\text{LO}}_{R};
         &\dot{\vec{q}}^+ = \Delta_{\mathcal{F}_{R} \rightarrow \mathcal{F}_{T}} \dot{\vec{q}}^-,
         &\hspace{-0.2cm}\left( \vec{q},\dot{\vec{q}} \right)  \in \mathcal{S}^{\text{LO}}_{R}; \\
         \mathcal{F}_T,
         &\hspace{-0.2cm}\left( \vec{q},\dot{\vec{q}} \right) \not \in \mathcal{S}^{\text{TD}}_{F};
         &\dot{\vec{q}}^+ = \Delta_{\mathcal{F}_{T} \rightarrow \mathcal{F}_{F}} \dot{\vec{q}}^-,
         &\hspace{-0.2cm}\left( \vec{q},\dot{\vec{q}} \right)  \in \mathcal{S}^{\text{TD}}_{F}; \\
    \end{array}
\right .$} 
\end{equation}   

\begin{equation} 
\label{eq:HybridModelBG}
\resizebox{.89 \linewidth}{!}{$ \hspace{-0.2cm}\Sigma_{\text{BG}}:    \left\{ 
    \begin{array}{rrll}
         \mathcal{F}_F,
         &\hspace{-0.2cm}\left( \vec{q},\dot{\vec{q}} \right) \not \in \mathcal{S}^{\text{LO}}_{F};
         &\dot{\vec{q}}^+ = \Delta_{\mathcal{F}_{F} \rightarrow \mathcal{F}_{T}} \dot{\vec{q}}^-,
         &\hspace{-0.2cm}\left( \vec{q},\dot{\vec{q}} \right) \in \mathcal{S}^{\text{LO}}_{F}; \\
         \mathcal{F}_T,
         &\hspace{-0.2cm}\left( \vec{q},\dot{\vec{q}} \right) \not \in \mathcal{S}^{\text{TD}}_{R};
         &\dot{\vec{q}}^+ = \Delta_{\mathcal{F}_{T} \rightarrow \mathcal{F}_{R}} \dot{\vec{q}}^-,
         &\hspace{-0.2cm}\left( \vec{q},\dot{\vec{q}} \right)  \in \mathcal{S}^{\text{TD}}_{R}; \\
         \mathcal{F}_R,
         &\hspace{-0.2cm}\left( \vec{q},\dot{\vec{q}} \right) \not \in \mathcal{S}^{\text{TD}}_{F};
         &\dot{\vec{q}}^+ = \Delta_{\mathcal{F}_{R} \rightarrow \mathcal{F}_{A}} \dot{\vec{q}}^-,
         &\hspace{-0.2cm}\left( \vec{q},\dot{\vec{q}} \right)  \in \mathcal{S}^{\text{TD}}_{F}; \\
         \mathcal{F}_A,
         &\hspace{-0.2cm}\left( \vec{q},\dot{\vec{q}} \right) \not \in \mathcal{S}^{\text{LO}}_{R};
         &\dot{\vec{q}}^+ = \Delta_{\mathcal{F}_{A} \rightarrow \mathcal{F}_{F}} \dot{\vec{q}}^-,
         &\hspace{-0.2cm}\left( \vec{q},\dot{\vec{q}} \right)  \in \mathcal{S}^{\text{LO}}_{R}; \\
    \end{array}
\right .$} 
\end{equation}    

\begin{equation} 
\resizebox{.89 \linewidth}{!}{$ \hspace{-0.2cm} \Sigma_{\text{B2}}: \left\{ 
    \begin{array}{rrll}
         \mathcal{F}_F,
         &\hspace{-0.2cm}\left( \vec{q},\dot{\vec{q}} \right) \not \in \mathcal{S}^{\text{LO}}_{F};
         &\dot{\vec{q}}^+ = \Delta_{\mathcal{F}_{F} \rightarrow \mathcal{F}_{T}} \dot{\vec{q}}^-,
         &\hspace{-0.2cm}\left( \vec{q},\dot{\vec{q}} \right) \in \mathcal{S}^{\text{LO}}_{F}; \\
         \mathcal{F}_T,
         &\hspace{-0.2cm}\left( \vec{q},\dot{\vec{q}} \right) \not \in \mathcal{S}^{\text{TD}}_{R};
         &\dot{\vec{q}}^+ = \Delta_{\mathcal{F}_{T} \rightarrow \mathcal{F}_{R}} \dot{\vec{q}}^-,
         &\hspace{-0.2cm}\left( \vec{q},\dot{\vec{q}} \right)  \in \mathcal{S}^{\text{TD}}_{R}; \\
         \mathcal{F}_R,
         &\hspace{-0.2cm}\left( \vec{q},\dot{\vec{q}} \right) \not \in \mathcal{S}^{\text{LO}}_{R};
         &\dot{\vec{q}}^+ = \Delta_{\mathcal{F}_{R} \rightarrow \mathcal{F}_{T}} \dot{\vec{q}}^-,
         &\hspace{-0.2cm}\left( \vec{q},\dot{\vec{q}} \right)  \in \mathcal{S}^{\text{LO}}_{R}; \\
         \mathcal{F}_T,
         &\hspace{-0.2cm}\left( \vec{q},\dot{\vec{q}} \right) \not \in \mathcal{S}^{\text{TD}}_{F};
         &\dot{\vec{q}}^+ = \Delta_{\mathcal{F}_{T} \rightarrow \mathcal{F}_{F}} \dot{\vec{q}}^-,
         &\hspace{-0.2cm}\left( \vec{q},\dot{\vec{q}} \right)  \in \mathcal{S}^{\text{TD}}_{F}; \\
    \end{array}
    \right .$}    
\label{eq:HybridModelB2}
\end{equation}    

\subsection{Finding Optimal Gaits using Trajectory Optimizations}
There are infinitely many periodic solutions of the aforementioned hybrid models.
In order to compare the underlying dynamics and understand the motion characteristics of the pronking and bounding gaits, we seek to identify the optimal periodic solutions across a range of speeds.
This gait generation task can be formulated as a trajectory optimization (TO) problem.
Specifically, finding a gait of the robot is equivalent to finding states trajectories to one of the hybrid models (\crefrange{eq:HybridModelPronk}{eq:HybridModelB2}) while subject to the following set of constraints:

\vspace{2mm}
\noindent\fbox{%
    \resizebox{.94 \linewidth}{!}{
      \parbox{\linewidth}{%
        \begin{itemize}
            \item Dynamics constraints (\ref{eq:DAE}) 
            \item Average sagittal speed: $\bar{v}_x = \frac{q_{x}(t_f) - q_{x}(t_0)}{t_f - t_0}$ m/s;
            \item Average lateral speed: $\bar{v}_y = \frac{q_{y}(t_f) - q_{y}(t_0)}{t_f - t_0} = 0$ m/s;
            \item Ground clearance constraints in flight: $p^z_i(\vec{q}) > 0$ ;
            \item Periodicity constraints: $\vec{q}(t_0) = \vec{q}(t_f)$, $q_{x}(t_0) \ne q_{x}(t_f)$;
            \item Configuration limits: $\vec{q}_{\text{min}}\leq\vec{q}\leq\vec{q}_{\text{max}}$;
            \item Velocity limits: $|\dot{\vec{q}}|\:\leq\dot{\vec{q}}_{\text{max}}$;
            \item Torque limits: $|\vec{\tau}|\:\leq\vec{\tau}_{\text{max}}$;
            \item Friction cone limits: $ \mu \: < 0.6$;
            \item Swing foot linear and angular velocities at touch-down;
        \end{itemize}
      }
    }
}

\vspace{1mm}
Furthermore, a virtual constraint is added for each joint to follow a 5-th order B\'ezier polynomial:
\begin{equation}
\label{eq:bezier ploynomials}
\resizebox{.6 \linewidth}{!}{$
  h(s) =    \sum_{r=0}^{5} {\scriptstyle  \dfrac{5 ! \alpha_{r}}{r !(5-r) !} } s^{r}(1-s)^{5-r}
$}
\end{equation}
%
where $s \in [0,1]$ is the normalized phase time and $\alpha_r$ are the constant B\'ezier coefficients.
As a result, each domain of an optimal solution can be fully represented by a numerical matrix $\vec{B}(\bar{v}_x)  \in \R^{12 \times \left(5+1\right)}$ parameterized by the average velocity $\bar{v}_x$ and contains the set of coefficients $\alpha_{r}$ of the B\'ezier polynomials for all joints.
In order to ensure the bilateral symmetry of legs, we also introduce virtual constraints by equating the coefficients of the B\'ezier polynomials for the joints on both sides.

The solutions are optimized to obtain the desired gait at various average speeds while minimizing the cost of transport (COT):
\begin{equation}
\label{eq:cost function}
    \text{COT} =\frac{\int_{t_o}^{t_f} \sum_{k=1}^{12} | \tau_k \cdot \dot{q}_{\text{leg},k}| \:d t}{mg \left[ \: q_{x} {\scriptstyle (t_f)} - q_{x}{\scriptstyle (t_o)} \: \right]}  
\end{equation}
where $t_o$ and $t_f$ are the start and end times of a full stride, and $\tau_k$ and $\dot{q}_{\text{leg},k}$ are the torque and the velocity of the $k$-th joint. 
$mg$ is the total weight of the robot and $ q_{x} {\scriptstyle (t_f)} - q_{x}{\scriptstyle (t_o)}$ is the distance traveled in $x$ direction. The absolute value operator is used to include the positive and negative work.

The NLP optimization problem were constructed and solved using the open-source framework FROST developed by Hereid \etal \cite{hereid2017frost, hereid2019rapid}.

\section{Results}
\label{sec:results}
\begin{figure}[t!]
\centering
\includegraphics[width=1\columnwidth]{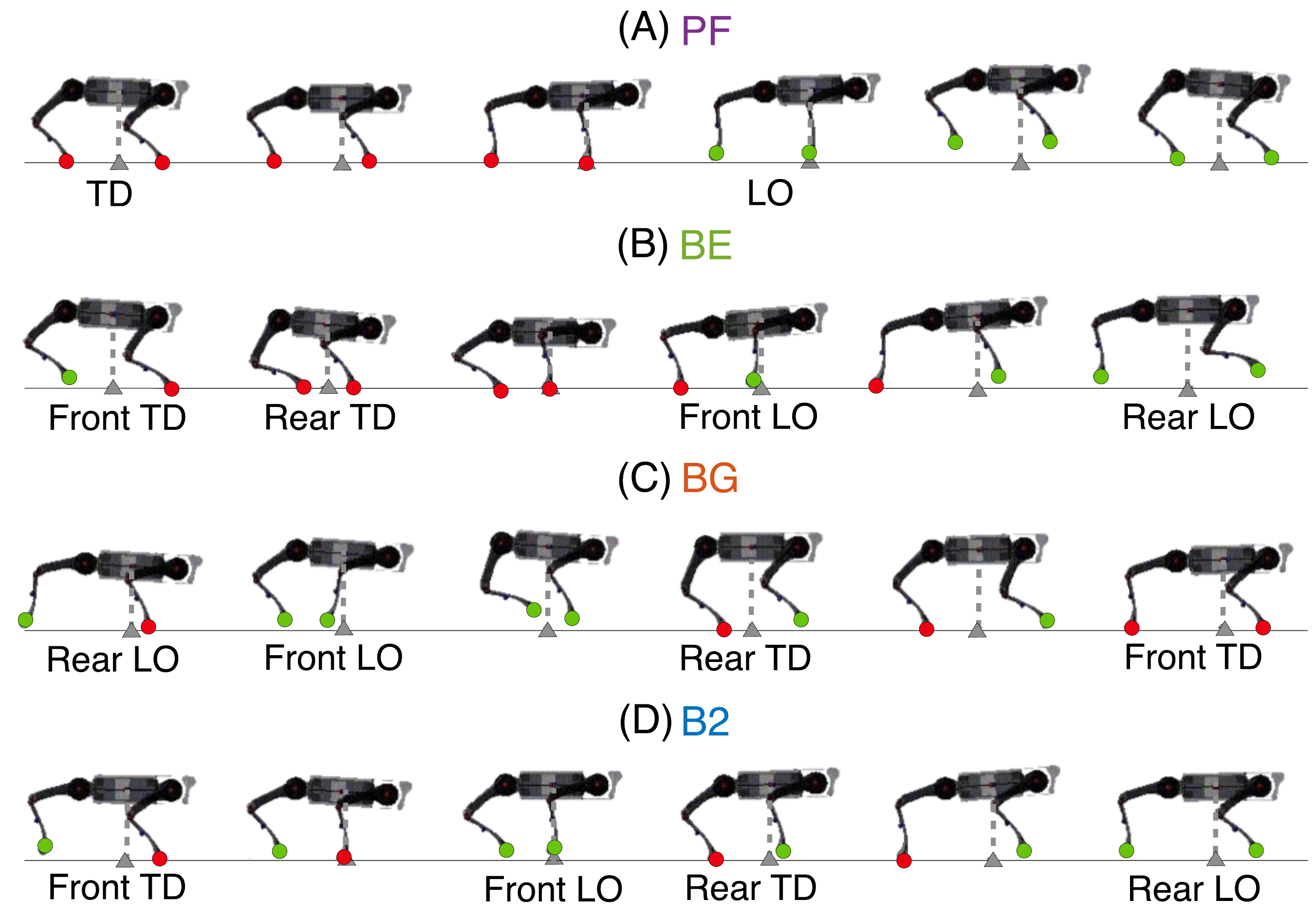}
\caption[COT]{A succession of key frames illustrates the optimal motion for each of the four gaits at a speed of 3 m/s. From the top to the bottom, they are (A) pronking forward, (B) bounding with extended suspension, (C) bounding with gathered suspension, and (D) bounding with two flight phases.}
\label{fig:OptimalGaits}
\vspace{-2mm}
\end{figure}
This section is divided into three parts.
A discussion of the cost of transport for each of the four gaits at different speeds was presented in the first part.
After that, we continued to examine the positive and negative work performed by the thigh and calf joints in both the front and rear leg pairs for the three bounding gaits. Furthermore, we compared joints work to the stride time and duty factor to provide insight into the legs' behavior. The final part focused on the exchange of angular momentum between the torso and the legs during different phases in the three bounding gaits.    
\subsection{Identified Optimal Gaits}
\label{sec:GaitLibrary}

Throughout the study, we examined locomotion velocities between 0 m/s and 5 m/s with an increment of 0.1 m/s for each gait.
Our first optimization was performed at a speed of 2 m/s using an iterative optimization method developed in \cite{yesilevskiy2018spine}.
The following optimizations were performed at neighboring speeds until the whole range of speeds was covered, or no solution could be identified.
Each optimization of a bounding gait took an average of 800 iterations or 15 minutes on a desktop computer to converge.
As the optimization problem was re-initialized and recalculated multiple times at each speed, finding an optimal gait took 26 minutes on average. 
We were unable to find optimal solutions when locomotion speeds approached 5 m/s or were close to zero due to the joint angle and torque limits.
Optimal pronking gaits were identified with average speeds ranging from 0.1 m/s to 4.7 m/s.
The minimum average speed of BG is 0.4 m/s, and 4.4 m/s is the maximum.
Among the three bounding gaits, BE has the shortest range of speed between 0.5 m/s and 3.6 m/s. 
By contrast, B2 has the largest range of speeds, ranging from 0.3 to 4.6 m/s.
Snapshots of video animations (created using optimal solutions) of a full stride for all four gaits at a speed of 3 m/s have been demonstrated and compared in Fig~\ref{fig:OptimalGaits}.



\begin{figure}[t!]
\centering
\includegraphics[width=1\columnwidth]{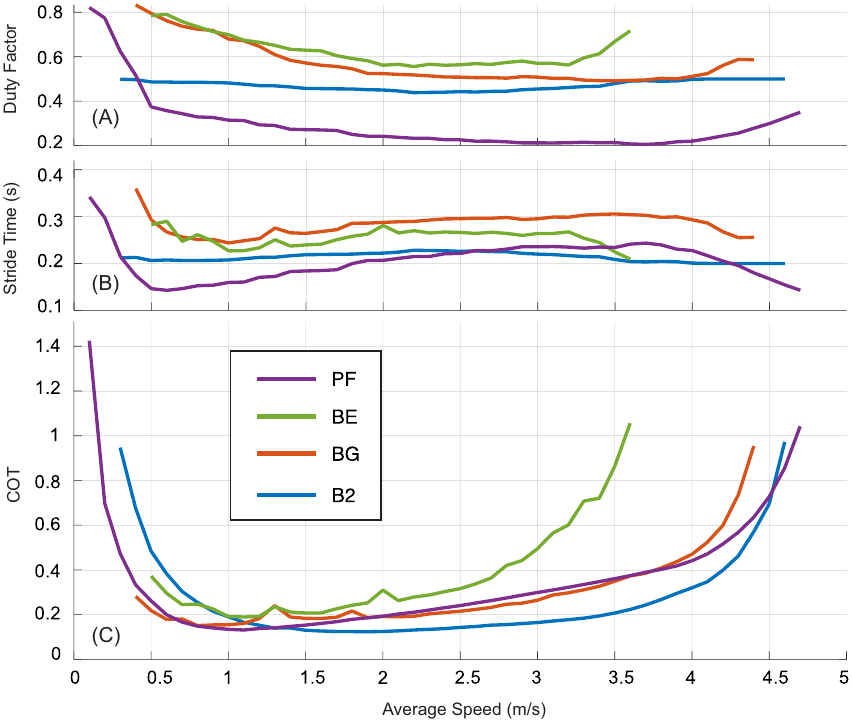}
\caption[COT]{This figure shows the duty factor, the total stride time, and the cost of transport as functions of forward speed for pronking forward (PF) and the three bounding gaits (BE, BG, and B2). Among gaits moving between 1.3 m/s and 4.5 m/s, we found the bounding gait with two flight phases (B2) to be the most energy efficient.}
\label{fig:COT}
\vspace{-2mm}
\end{figure}


Fig.~\ref{fig:COT} shows the duty factor (the percentage of stance phases time in a full stride) \cite{Alexander2002PrincipleofLocomotion}, the total stride time, and COT of the optimized solutions for each gait following the style of efficient gait study of horses \cite{hoyt1981gait}.
Each curve in Fig.~\ref{fig:COT}(C) represents A1's COT values as a function of average forward speed. 
At the speed of 0.1 m/s, the only feasible gait for A1 is pronking forward gait (PF, purple curves in Fig.~\ref{fig:COT}).
A stride time of 0.34 seconds is achieved at this speed, and the duty factor at this speed is 0.82, resulting in a very inefficient gait. 
The bounding gait with two flight phases (B2, blue curves in Fig.~\ref{fig:COT}) emerges at 0.3 m/s, with a duty factor about 0.5 and a stride time of 0.21 s.
It is a gait with a very high energy cost for the A1 robot and is less efficient than the PF gait at relatively low speeds.
Between 0.4 m/s and 0.6 m/s, bounding with gathered suspension (BG, red curves) appears and becomes the most energetically efficient gait, while B2 remains the most energy-intensive gait, with a COT value 50\% higher than others.
As speed further increases, PF is slightly more efficient than BG, and it reaches the optimal speed at 1.1 m/s with the smallest COT value of 0.33.
When forward speed exceeds 1.3 m/s, B2 surpasses PF and becomes the most efficient gait up to 4.5 m/s.
In addition, bounding with extended suspension (BE, green curves) is almost always the most energy-intensive gait across the whole range of speeds.
Further, it is intriguing to note that among all four gaits, the duty factor and stride time of B2 are almost constant at 0.5 and 0.2 s, respectively, whereas they vary rapidly for the other gaits, especially at low and high speeds.

\subsection {Joint Work}
\begin{figure}[tbp]
\centering
\includegraphics[width=1\columnwidth]{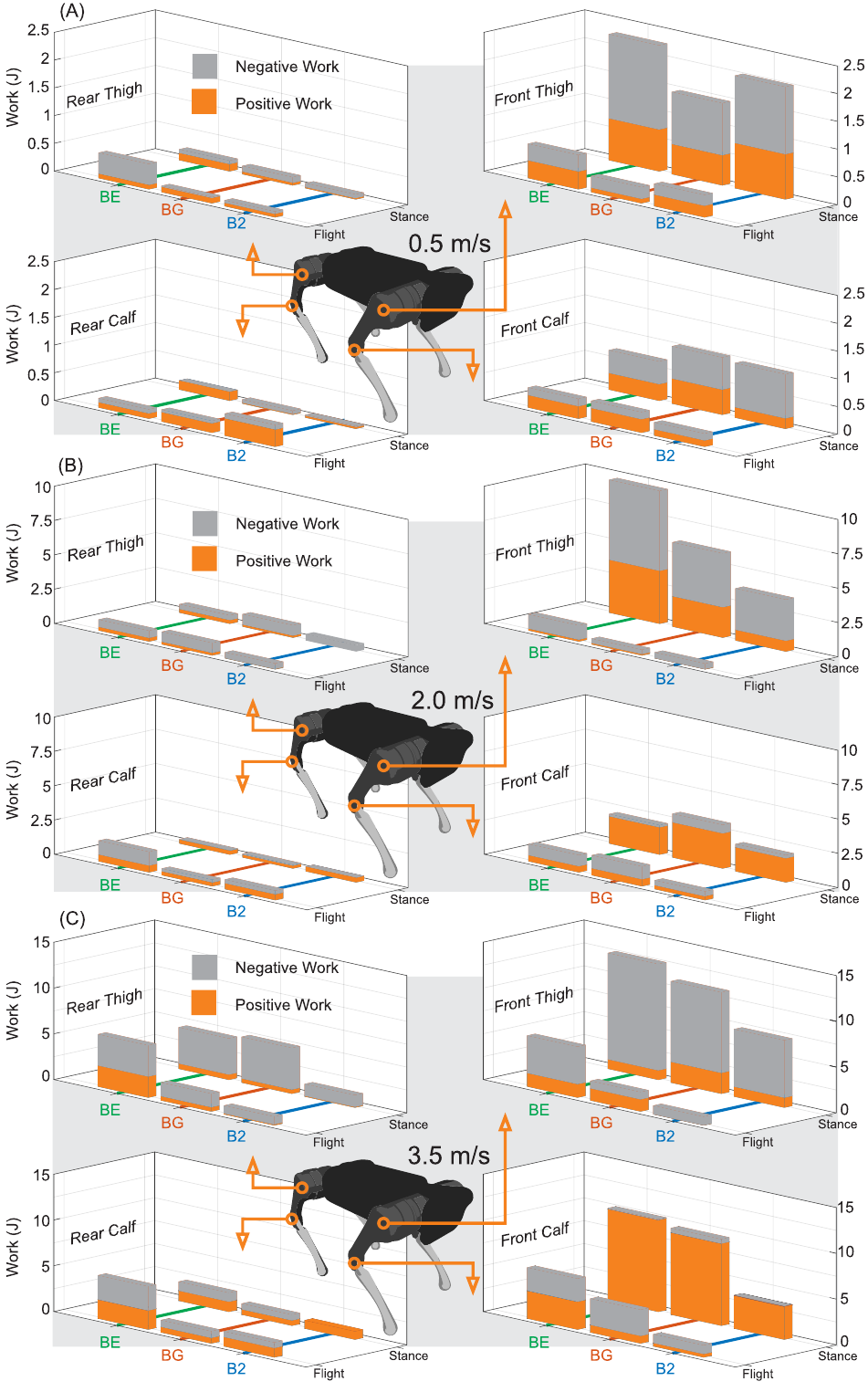}
\caption[Joints Work]{This figure shows the positive and negative work performed by each joint within the whole stride of BE, BG, and B2 at the speed of 0.5 m/s, 2.0 m/s, and 3.5 m/s, respectively.}
\label{fig:JointWork}
\vspace{-2mm}
\end{figure}

The work performed by each joint during both the stance and flight phases was investigated to gain a better understanding of the energy consumption of the three bounding gaits.
The hip joints' motions were excluded from this analysis since the motion is only considered in the sagittal plane, and the hip joints were constrained to be stationary for each gait.
This analysis was conducted at three different speeds: 0.5 m/s, 2.0 m/s, and 3.5 m/s.
The bar graphs in Fig.~\ref{fig:JointWork} illustrate the work performed in the rear thigh, front thigh, rear calf, and front calf joints during a full stride.
As legs moved in pairs, only the right sides of the joints were demonstrated.
Additionally, positive work (orange bar) and negative work (grey bar) have been distinguished when the leg is in the swing or stance phase.
When a joint has a positive work, it means that the actuator has produced a torque acting in the same direction as the joint rotation, which adds energy to the system (speeding up). On the other hand, joints that have negative work indicate that the motors are dissipating the total energy of the system (braking or slowing down).
\label{sec:Total Work}

\subsubsection{Low Speeds (0.5 m/s)} we observed similar energy consumption in the three bounding gaits when the locomotion speed was low.
Bounding gaits do not maintain symmetries in motion between the front and rear legs, causing the legs to behave differently than symmetrical gaits such as trotting \cite{hildebrand1965symmetrical}.
There was almost equal contribution between the positive and negative work of the front leg pairs during the stance phase in all three gaits: during the first half of the stance phase, they were mostly performing negative work to slow down the system, and after mid-stance, they accelerated the motion and restored energy to the system.
By contrast, the joints in the hind legs required significantly less effort since the torso vaulted over them as if they were inverted pendulums.
These motions are illustrated in Fig.~\ref{fig:OptimalGaits}.

\subsubsection{Intermediate Speeds (2.0 m/s)} as the A1 robot approached mid-speed, the leg pairs exhibited similar behavior as they did while bounding at low speeds.
In all three bounding gaits, it has been demonstrated in Fig.~\ref{fig:JointWork}(B), although the total amount of work performed by all the joints increased with speed, it was primarily the front thigh joints that contributed the most to the total work to propel the robot forward, while other joints played a relatively small role.
There was, however, a trend in bounding gaits in which the front calf joints performed more positive work, and the front thigh joints braked more frequently with increasing speed.
The front thigh joint performed 3.8 J of work during stance in B2, which was only 1.8 J more than the front thigh joint's work at 0.5 m/s.
Comparatively, the energy expenditure in the other two bounding gaits was greater.
Just during the stance phase, the front thigh joints experienced an energy increase of 4.7 J and 7.4 J in BG and BE, respectively.
\subsubsection{High Speeds (3.5 m/s)} when the legs were in swing, the amount of work performed in all joints increased significantly as locomotion speed increased to top speeds.
Especially in BE, the energy expenditure was almost doubled in all joints. Additionally, in the stance phase, it became apparent that the front thigh and front calf joints had distinct functionalities: the front thigh joints almost exclusively contributed to the negative work, while the front calf joints largely contributed to the positive work.
In the front calf joints alone, the positive work increased by 8.1 J for BE and 6.4 J for BG in comparison with the energy expenditure at the speed of 2 m/s in the same joint.
It is important to note, however, that the positive work performed by the calves of the front leg in B2 at the top speed was only one third of that of BE and BG at the same speed.
One explanation is that because the torso was modeled as a rigid body with a constant moment of inertia, the natural oscillation frequency for pitching rotation is fixed.
However, the total stride time and duty factors for BE and BG changed rapidly, while these values remained relatively unchanged for B2 as shown in Fig.~\ref{fig:COT}(A)\&(B).
As stride frequencies $f = \nicefrac{1}{T}$ moved away from their natural oscillation frequencies, the primary function of front and rear thigh joints was to apply torques to modulate torso's oscillation in order to maintain the required footfall patterns.
Only B2 maintained its stride frequency at its optimal value (around 5 Hz), which made it the most efficient gait.

\begin{figure}[tbp]
\centering
\includegraphics[width=1\columnwidth]{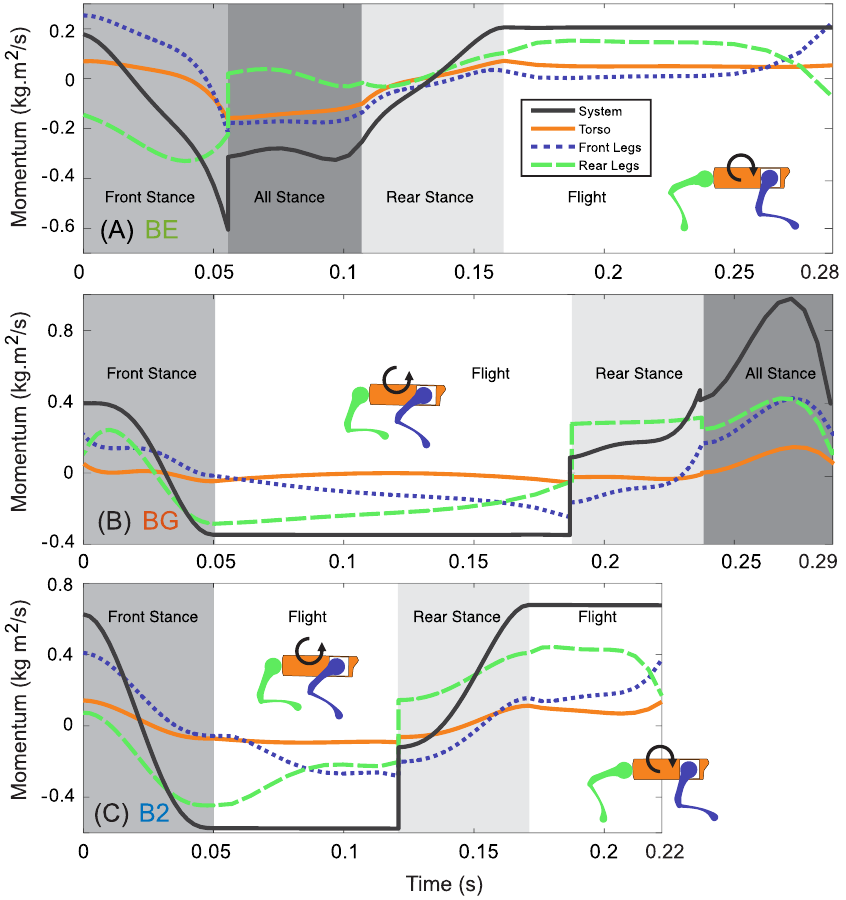}
\caption[Momentum Exchange]{This figure depicts the exchange of angular momentum in three bounding gaits BE, BG, and B2 during the entire stride at a speed of 2 m/s.}
\label{fig:AngularMomentum}
\vspace{-2mm}
\end{figure}

\subsection {Exchange of Angular Momentum}
In bounding gaits, the leg pairs strike the ground at different timings, which inevitably causes the torso to rotate throughout the gait cycle. 
Since the legs of A1 contribute about 60\% of the robot's total mass, it is also imperative to examine the effect of the legs swinging motion to comprehend the dynamics of optimal bounding gaits with different footfall patterns.
In this section, we illustrated the angular momentum about the center of mass (COM) of the whole system as a function of time for all three bounding gaits at aspeed 2 m/s in Fig.~\ref{fig:AngularMomentum}.
To better understand the exchange of angular momentum between the system components, we separated the total angular momentum of the system (solid black curves) from the angular momentum of the torso (solid orange curves), the front legs (dotted blue curves), and the rear legs (dashed green curves).
Because the motion of the robot was restrained to the $x{\text -}z$ plane, only the angular momentum about the $y$ axis (pitching) was discussed.
Positive values in the figure indicate angular momentum in the clockwise direction.

Due to the conservation of the angular momentum, in all flight phases, the angular momentum of the whole system was a constant value.
Also, during flight, the front and rear legs exchanged angular momentum by the same amount, while the angular momentum of the torso remained almost unchanged.
The total angular momentum in BG and B2 was negative after the front stance, resulting in a counterclockwise rotation of the torso during the flight phase.
On the other hand, the total angular momentum in BE and B2 was positive after the rear leg stance, which caused the torso to rotate clockwise.
As the system entered the stance phase, the angular momentum of the system began to change rapidly.
Regardless of differences in footfall patterns, all three bounding gaits displayed similar leg behaviors:
Angular momentum changed from positive to negative during the front leg stance phase, and the reverse occurred during the rear leg stance phase with an increase in angular momentum in all components.
Additionally, in all three gaits we noticed there were large jumps in the angular momentum for the rear legs at the moment touch-down, while the front legs almost matched the ground speed and the changes in the angular momentum were almost negligible at the ground contact.
A comparison of the angular momentum over the entire stride in all three gaits revealed that the amount of total angular momentum in B2 during the two flight phases was greater than that in BE and BG in similar phases.
During B2, the two leg pairs can effectively alter the angular momentum by applying forces in the two separated stance phases without interfering with each other.
In contrast, the largest angular momentum in BE was only about 0.2 $kg \cdot m^2/s$.
Because in the all stance phase, all four feet were very close to each other, as shown in Fig.~\ref{fig:OptimalGaits}(B), the motors have difficulty redirecting the linear motion of the torso while adjusting angular momentum.


\section{Conclusions}
In this paper, we investigated the energetics of the pronking gait and three types of bounding gaits for the quadrupedal robot A1.
We created a realistic full-body model for the robot and formulated each gait as a hybrid system.
Then, we conducted trajectory optimizations and used the FROST framework to minimize the cost of transport between the speed of 0 m/s and 5 m/s.
Furthermore, we examined the positive and negative work done by each joint in the robot during the whole stride cycle for all bounding gaits at three velocities.
Lastly, we illustrated the exchange of the angular momentum between the torso, front legs, and rear legs of the robot at speed 2 m/s to reveal some insights into the underlying dynamics.  

In general, it has been found that bounding with two flight phases (B2) is the most energy efficient gait for A1. 
At a wide range of speeds, this gait exhibits a nearly constant duty factor and stride time of 0.5 and 0.2 seconds, respectively. 
Our examination of the movements of the joints also revealed that the front legs were primarily used to propel the robot forward during all three bounding gaits, particularly at relatively low speeds.
However, the hind legs usually struck the ground rigidly during the rear stance phase, resulting in considerable angular momentum losses.
As shown in Fig.~\ref{fig:Kinematics Diagram}(A), the A1 robot has an asymmetrical design with all the calf joints inverted backward.
Consequently, the COM of the entire system has been shifted towards the edge of the support polygon at the moment of front legs lift-off as seen in Fig.~\ref{fig:OptimalGaits}. We assume that the uneven distribution of limb forces during motion is caused by the structural design of the robot. Compared with symmetrical gaits such as trotting, this phenomenon is exacerbated by bounding gaits, since phase delays between the two leg pairs lead to more pronounced torso rotations.

To validate this assumption, in future work, we can potentially vary the location of the COM for the torso in our model and investigate how that will affect forces on the limbs and the optimal bounding gaits.
Alternatively, we can rerun the trajectory optimizations and command the robot to move backward.
It is equivalent to change the legs configuration in the system by bending the calf joints in the opposite direction.
We should expect drastic changes in the energetics of the three bounding gaits.

\label{sec:conclusions}

\bibliographystyle{IEEEtran}
\bibliography{References}

\end{document}